\documentclass{article}

\usepackage{microtype}
\usepackage{appendix}
\usepackage{graphicx}
\usepackage{booktabs} 
\usepackage{MnSymbol,wasysym}
\usepackage{pifont}
\usepackage{etoolbox}
\usepackage{siunitx}
\usepackage{enumitem}
\usepackage{caption}
\usepackage[small,compact]{titlesec}
\usepackage{amsmath}
\usepackage{algorithmic}
\usepackage{textcomp}
\usepackage{xcolor}
\usepackage{lipsum}
\usepackage[first=1,last=100]{lcg}
\usepackage{blindtext}
\usepackage{adjustbox}
\usepackage{multicol}

\usepackage{hyperref}

\usepackage[accepted]{mlsys2022}

\usepackage{url}

\begin{document}

\twocolumn[
\mlsystitle{Machine Learning Sensors
}

\raggedbottom



\mlsyssetsymbol{equal}{*}

\begin{mlsysauthorlist}
\mlsysauthor{Pete Warden}{stanford}
\mlsysauthor{Matthew Stewart}{harvard}
\mlsysauthor{Brian Plancher}{harvard}
\mlsysauthor{Colby Banbury}{harvard}
\mlsysauthor{Shvetank Prakash}{harvard}
\mlsysauthor{Emma Chen}{harvard}
\mlsysauthor{Zain Asgar}{stanford}
\mlsysauthor{Sachin Katti}{stanford}
\mlsysauthor{Vijay Janapa Reddi}{harvard}\\\vspace{1em}
$^1$Stanford University~~~$^2$Harvard University

\end{mlsysauthorlist}
\mlsysaffiliation{harvard}{Harvard University}
\mlsysaffiliation{stanford}{Stanford University}

\mlsyskeywords{Machine Learning, MLSys}

\vspace*{1em}

\begin{abstract}

Machine learning sensors represent a paradigm shift for the future of embedded machine learning applications. 
Current instantiations of embedded machine learning (ML) suffer from complex integration, lack of modularity, and privacy and security concerns from data movement.
This article proposes a more data-centric paradigm for embedding sensor intelligence on edge devices to combat these challenges.
Our vision for ``sensor 2.0'' entails segregating sensor input data and ML processing from the wider system at the hardware level and providing a thin interface that mimics traditional sensors in functionality. 
This separation leads to a modular and easy-to-use ML sensor device.
We discuss challenges presented by the standard approach of building ML processing into the software stack of the controlling microprocessor on an embedded system and how the modularity of ML sensors alleviates these problems.
ML sensors increase privacy and accuracy while making it easier for system builders to integrate ML into their products as a simple component.
We provide examples of prospective ML sensors and an illustrative datasheet as a demonstration and hope that this will build a dialogue to progress us towards sensor 2.0.

\end{abstract}

]




\section{Introduction}
\label{sec:introduction}

Since the advent of AlexNet~\cite{NIPS2012_c399862d}, deep neural networks have proven to be robust solutions to many challenges that involve making sense of data from the physical world. Machine learning (ML) models can now run on low-cost, low-power hardware capable of deployment as part of an embedded device. Processing data close to the sensor on an embedded device allows for an expansive new variety of always-on ML use-cases that preserve bandwidth, latency, and energy while improving responsiveness and maintaining data privacy. 
This emerging field, commonly referred to as embedded ML or tiny machine learning (TinyML)~\cite{warden2019tinyml,david2021tensorflow,reddi2021widening,TinyMach84:online}, is paving the way for a prosperous new array of use-cases, from personalized health initiatives to improving manufacturing productivity and everything in-between.

However, the current practice for combining inference and sensing is cumbersome and raises the barrier of entry to embedded ML. 
At present, the general design practice is to design or leverage a board with decoupled sensors and compute (in the form of a microcontroller or DSP), and for the developer to figure out how to run ML on these embedded platforms. The developer is expected to train and optimize ML models 
and fit them
within the resource constraints of the embedded device. Once an acceptable prototype implementation is developed, the model is integrated with the rest of the software on the device. Finally, the widget is tethered to the device under test to run inference.%
\begin{figure}[!h]
\centering
    {\includegraphics[trim=175 240 200 25, clip, width=\columnwidth]{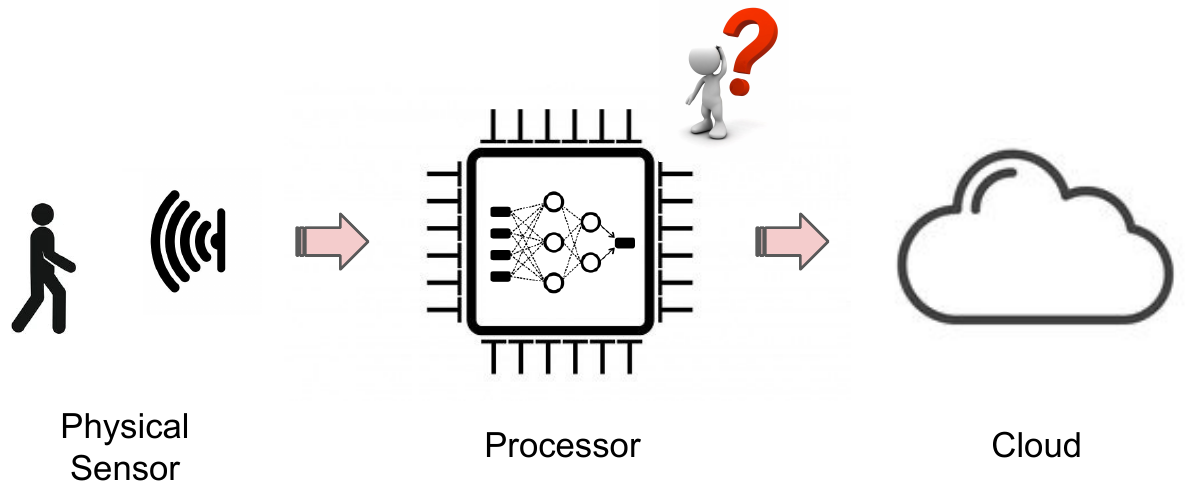}}
    \caption{The Sensor 1.0 paradigm tightly couples the ML model with the application processor and logic, making it difficult to provide hard guarantees about the ML sensor's ultimate behavior.}
\label{fig:ml_sensor_v1}
\end{figure}
\begin{figure}[!h]
\centering
    {\includegraphics[trim=100 60 200 170, clip, width=\columnwidth]{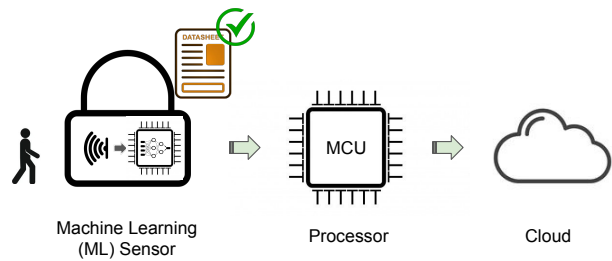}}
    \caption{Our proposed Sensor 2.0 paradigm. The ML model is tightly coupled with the physical sensor, separate from the application processor, and comes with an ML sensor datasheet that makes its behavior transparent to the system integrators and developers.}
    \vspace{-13pt}
\label{fig:ml_sensor_v2}
\end{figure}
%
The current approach is slow,
manual, energy-inefficient, and error-prone. It requires a sophisticated understanding of ML and the intricacies of ML model implementations to optimize and fit a model within the constraints of the embedded device.
\newpage
To unlock the full potential of ML sensors, we must usher in a new way of thinking about ML and sensing by raising the level of abstraction for ML sensors from the current approach (Figure~\ref{fig:ml_sensor_v1}) to a self-contained (hardware and software) system that utilizes on-device ML to extract useful information from some complex set of phenomena in the physical world and reporting it through a simple interface to a wider system (Figure~\ref{fig:ml_sensor_v2}). For example, consider an ML sensor designed for person detection~\cite{tflitemi78:online}. Such a device is often used to wake up a user interface and automatically adjust environmental controls like lighting, air conditioning, and heating. In our ``Sensor 2.0'' paradigm, such a sensor would have a very minimal interface with an opening that points outward at the environment and only three external pins: two for power and ground, and one signal pin that is driven high when a person is detected and held high for as long as people are present. 
As this information should be all the knowledge that a future ML sensor system engineer needs to know, such a system would be ready to use out-of-the-box, easing practical deployment.

The above approach has the added benefit of providing more granular control on data and how it is consumed. By constraining the interface between the ML sensor and the rest of the software to be around inference outcomes and not allowing access to raw sensor data, we avoid the potential for privacy breaches, whether they are due to malicious intent or errors. We can also reason about the data flow in a principled manner since the ML sensor can now have a "data" datasheet, specifying what kind of data is exposed. This approach enables independent verification without trusting software from the developer. It opens new paths to building modular ML sensor-based systems for smart homes, enterprise IoT, and other domains. This method should also help inform discussions about the privacy and ethical issues of the technology, similar to how concerns were raised for traditional internet of things (IoT) devices~\cite{atlam2020iot,wolf2017safety,emami2020ask}.

ML sensors that adopt the sensor 2.0 paradigm are slowly starting to emerge, such as Qualcomm's Always-on Computer Vision Module (CVM)~\cite{Qualcomm86:online,Alwayson15:online} and Bosch's BHI260AP smart sensor~\cite{Smartsen25:online}. However, there is still a long way before the industry adopts a principled approach to ML sensor design. The current solutions are bespoke architectures that are proprietary and less than transparent. So we need to rethink how we systematically embed intelligence into sensor 2.0 devices. To this end, we present five tentative design principles to foster discussion within the community:

\begin{enumerate}
    \item We need to raise the right level of \textbf{abstraction} that will enable ease of use for scalable deployment of ML sensors; not everyone should be required to be an embedded systems developer or an engineer to use or leverage ML sensors into their ecosystem.
    \item The ML sensor's design should be inherently \textbf{data-centric} and defined by its input-output behavior instead of exposing the underlying hardware and software mechanisms that support ML model execution.
    \item An ML sensor's \textbf{implementation} must be clean and complexity-free. Features such as reusability, software updates, and networking must be thought through to ensure data privacy and secure execution.
    \item ML sensors must be \textbf{transparent}, indicating in a publicly and freely accessible ML sensor datasheet all the relevant information such as fact sheets, model cards, and dataset nutrition labels to supplement the traditional information available for hardware sensors.
    \item We as a community should aim to foster an \textbf{open ML sensors ecosystem} by maximizing data, model, and hardware transparency where possible, without necessarily relinquishing any claim to intellectual property.
\end{enumerate}

In summary, the overarching goal of our work is to focus new research and development efforts in a way that helps the field progress faster in delivering the potential benefits of the recent advances in ML and embedded hardware. To this end, Section~\ref{sec:challenges} outlines the challenges facing existing approaches for embedding ML intelligence into devices. Section~\ref{sec:approach} describes our proposed approach. Section~\ref{sec:datasheet} focuses on transparency of ML sensors via datasheets. Section~\ref{sec:ecosystem} talks about the ecosystem development around ML sensors. Section~\ref{sec:ethics} calls for ethical considerations around the pitfalls and dangers that we must address as ML sensors become ubiquitous. Section~\ref{sec:related} presents how ML sensors related to prior work. Finally,  Section~\ref{sec:conclusion} concludes the paper.
\section{Sensor 1.0 Challenges}
\label{sec:challenges}

\begin{figure}[!t]
    \includegraphics[trim = 30 0 50 0, clip, width=\columnwidth]{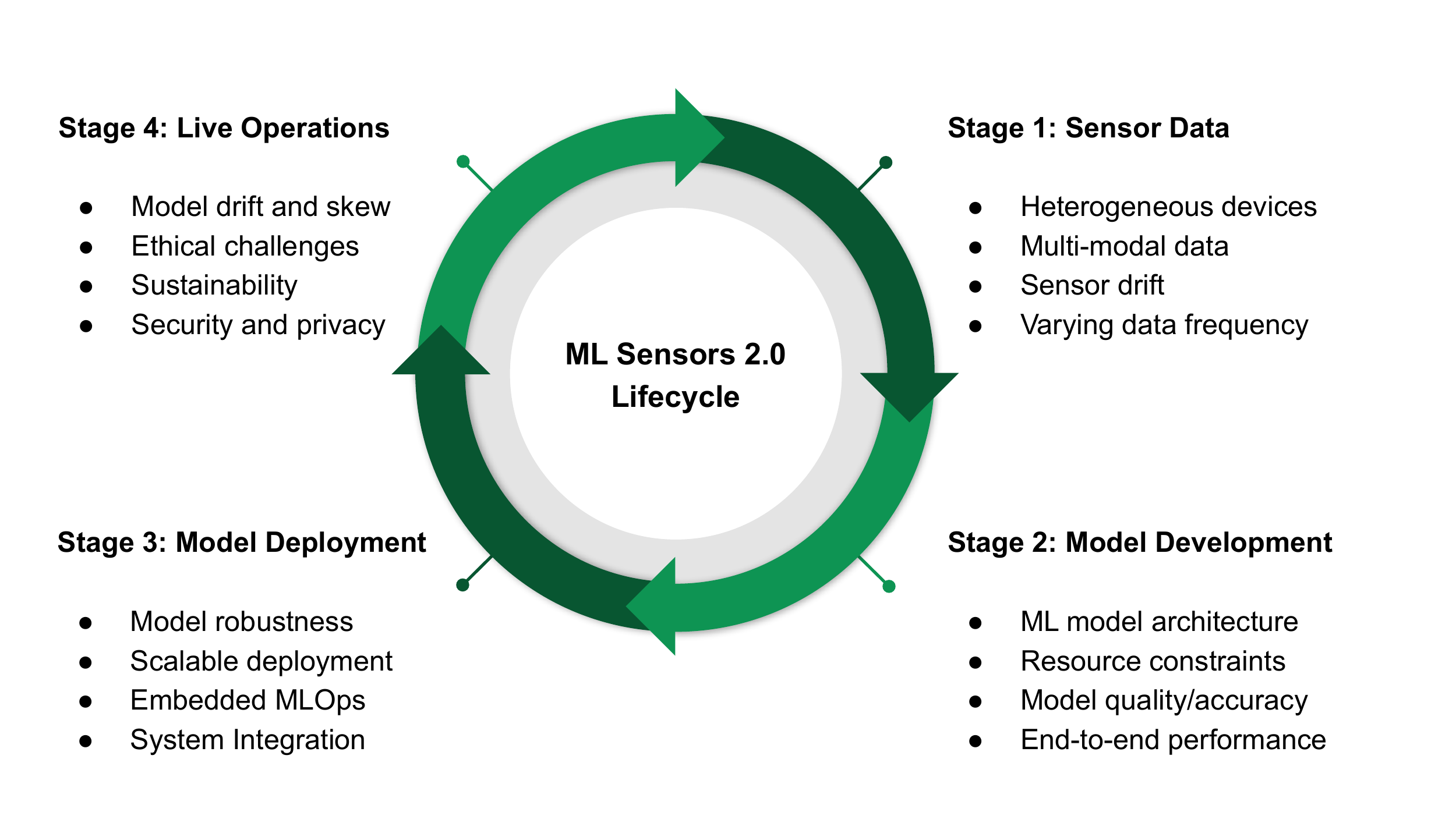}
    \caption{The machine learning lifecycle involves multiple stages of the pipeline, and several of these stages make the integration, development, and deployment of ML sensors extremely difficult.}
    \centering
    \label{fig:ml_lifecycle}
    \vspace{-10pt}
\end{figure}

This section discusses the challenges conventional approaches face in integrating and deploying sensor-based intelligence into embedded devices. While it is possible to launch successful embedded ML applications using a traditional low-level embedded systems approach, many challenges will likely occur during development. The resulting systems often have undesirable properties. We discuss these significant challenges in the context of the end-to-end major ML lifecycle stages, as illustrated in Figure~\ref{fig:ml_lifecycle}.

\subsection{Domain Skew Between Training and Sensor Data}

Data makes or breaks ML. The data set used for training an ML model is usually captured using different hardware and software than is used for a deployed system. These sorts of changes between the data between when you train and when you run inference is known as domain skew. Many properties can vary between sensors. For example, in the case of a sensor for person detection~\cite{chowdhery2019visual}, different implementations may have different camera modules, each with different input resolutions and sensitivity. 

Table~\ref{tab1} shows how sensor properties can vary across three commonly used embedded ML sensors. In general, if the properties are not the same between the sensors used to capture the training data and those in the deployed system, the model results are likely to be less accurate in production than an evaluation on the test set would indicate. This mismatch can contribute to compounding errors, known as ``data cascades,'' in the final application~\cite{sambasivan2021everyone}, problems that can be both hard to identify and costly to fix.

It is rare to capture all training data using the exact hardware and software processing pipeline present in the final system. Often the data used for training has been gathered by someone else and may not even include information on the characteristics relevant during live operations.

To compensate for these differences, it is first necessary to measure model accuracy when run with the final hardware and software configuration, feeding in representative captured data. This approach should give a reliable metric of the model’s accuracy in practice. If the accuracy is below what is required and expected from the training environment, steps need to be taken to reduce the differences between the training and production data. Such an approach might entail retraining with more representative data captured from the production environment, adjusting the sensor and processing to match the training data more closely, or combining the two approaches using synthetic data methods~\cite{7796926,nikolenko2021synthetic,Khaled}.

The domain skew becomes a barrier to deployment because particular sensors are often chosen for availability or cost reasons on a per-project basis. Hence, the compensation process has to be performed for each project too. Thus, skew can increase the cost of reusing models over time or unpredictably reduce model accuracy.

\subsection{Hardware \& Software Development Complexities}

Traditional software development is inherently different from software development for machine learning systems. These differences lead to a substantial ``AI Tax''~\cite{richins2021ai,buch2021ai} that can lead to overhead and complexity that is typically not present in traditional non-machine-learning systems. Managing these overheads and complexity implies we need first to understand these differences to address them.

Typical embedded software applications usually focus on signal processing, business logic, and hardware control. These areas are implemented using traditional procedural languages that typically involve data-conditional branching, scattering, and gathering and are memory-bound. Unlike these traditional workloads, machine-learning workloads typically involve hundreds of thousands of multiply-adds for each model inference, with highly-predictable access patterns, a low ratio of memory accesses to arithmetic operations, and very few data-dependent branches.

These differences require different trade-offs from hardware and system software designers to run the workloads effectively. For example, a data cache using a predictive algorithm will likely help speed up signal processing and business logic code. However, ML memory accesses can be accurately predicted at the compilation stage. So, controlling the schedule of loading memory into faster areas explicitly produces better results for those algorithms. 

\begin{table}[t!]
\caption{Varying Sensor Characteristics}
\begin{center}
\resizebox{\columnwidth}{!}{
\begin{tabular}{|c|c|c|}
\hline
\textbf{Microphones} & \textbf{Cameras}& \textbf{Accelerometers} \\ 
\hline
         Sampling rate & Field of View & Sampling Rate \\
         Digital or analog capturing & Shutter duration & Noise \\
         Echo cancellation & White Balance & Bias \\
         Noise Reduction & CCD Demosaicing & Drift \\
         Gain & Auto-exposure & \\
         Frequency Sensitivity & Resolution & \\
\hline
\end{tabular}
}
\vspace{-1em}
\label{tab1}
\end{center}
\end{table}

Another example is cooperative multitasking. Traditional embedded modules are unlikely to take over tens of thousands of cycles to complete an interaction since they execute hand-written logic or trigger hardware features and return, with the work often continued when an interrupt is signaled. So it is relatively easy to add cooperative multitasking handover calls at natural points in the code and have confidence that they will occur in a short and bounded interval. Because individual ML layers may take hundreds of thousands of cycles to complete and model invocations might take millions, it is not easy to hand over control at the expected frequency from the application logic. Instead, handover calls would need to be made from the inner loops of the layer implementations, which is much more invasive and requires preserving more states between yields.
\newpage
There are many more differences in the workloads, which has led to the development of specialized accelerator ASICs for ML in the embedded world and beyond. Even if such specialized accelerators are present, the embedded application still has to deal with extra complexity to orchestrate the model setup, invocations, and result handling, because these stages will require something more like a remote invocation over the bus than a regular branch to a function in the same address space. Any caches will need to be invalidated, for example, if shared memory has been altered.

Managing the ML computations imposes additional complexity on the embedded system's development, whatever the underlying implementation. This becomes a tax on development, as the hardware and operating system code designed for traditional embedded software struggle to handle the requirements of these new ML workloads.


\subsection{Testing and Debugging Difficulties for Deployment}

It is challenging to debug an embedded system when something goes wrong. The constraints on memory size, power, latency, and form factor make enforcing abstractions and modularity in the code difficult, and exposing state for observability is difficult. Their compute capabilities are also limited. These matters are complicated by adding ML to the mix since ML frameworks have large resource requirements.


Moreover, embedded applications are usually deployed as monolithic firmware binaries. Everything from the operating system and network stacks to the business logic is compiled into the same executable. More often than not, there are few accessible external boundaries where information can be gained about how the system works. Such a build environment makes testing individual modules hard because they often depend on other system parts. After all, the API abstractions are so leaky that extracting any single part and running it on a board or simulator will at least require compiling in the operating system and often other dependencies, too. Hence, it becomes impossible to test a module in isolation and have confidence in the results.

In practice, testing and debugging challenges arise due to the lack of unit tests. Tracking down the root cause of bugs takes much effort because they cannot be localized straightforwardly. From the authors' experience, even if a wake word model for keyword detection~\cite{https://doi.org/10.48550/arxiv.1711.07128,https://doi.org/10.48550/arxiv.1906.05721} is shipped as a library with minimal endpoints, a third party integrating the model may experience problems they believe are caused by the library code. The only feasible process to resolve the question of where the issues are is to reproduce the client's full executable and environment on a hardware simulator. Because the library and application share many resources, it is often unclear if a problem such as failing to process audio in the expected time is the fault of the framework code for 
\newpage
running too slowly or the client for not allowing enough free compute cycles for the library to run and execute. 

Similarly, given the limited memory protection, especially on ultra-low-power and low-cost embedded systems, corruption or illegal access faults can be hard to attribute to the right owners. There is a lack of enforcement mechanisms for module boundaries when running as part of a monolithic application, and this means that the contracts between libraries and clients are unclear. This is a challenge for ML frameworks because their functions tend to do more computation and resource allocation than other libraries typically used for embedded systems, so their requirements and constraints are also more complex.

\subsection{Sensor Data Auditability During Live Operation}

One of the significant concerns with deploying embedded devices is concern over the privacy of the users' data. Having an assurance that the raw data cannot be accessed and shared is vital for users to have confidence that they can safely deploy embedded devices into private spaces.

Cameras and microphones are ubiquitous input devices to ML models. These devices are often added to embedded systems to help answer questions like ``Is an object or person present?'' or ``Has a particular word been uttered or a sound heard?.'' The answers to these questions are usually not privacy-sensitive. They are stripped of unnecessary information such as an individual's identity or sounds and words that are not in the small subset within focus. In contrast, the raw camera and microphone outputs fed into the models have much more potential to be used maliciously.

The challenge is that the standard approach to deploying embedded models is integrating them as part of the same software system that handles all other aspects of the application (Figure~\ref{fig:ml_sensor_v1}). Consequently, raw sensor data is in an address space accessible by the rest of the code since embedded operating systems usually do not or cannot restrict access easily. To ensure that there are no deliberate or accidental back doors that enable cameras or microphones to be used as spying devices, it is necessary to audit all of the software on the system. Such an audit might include third-party network stacks that expose a large surface to remote attackers on a networked device. Even if there are no bugs, software updates may invalidate past audits and could be vectors for in-house attacks.


We need to understand how we isolate the raw data from leaking out to address this issue. Specifically, we need to architect the ML sensor from the hardware and software. Figure~\ref{fig:ml_sensor_v2} illustrates one possible way of ensuring separation. We elaborate on our proposal in greater detail in Section~\ref{sec:approach}.

\subsection{Full-Stack Skills for the Complete ML Lifecycle}

All of the above challenges are exacerbated by the software development approach to integrating ML models requiring a broad range of engineering skills beyond just understanding traditional embedded system development. This aspect makes deploying ML sensors extremely challenging. The skills gap must be bridged to address this concern as there is a dire need for infrastructure, frameworks, and tools that can help automate and streamline the processes.

Model authoring, data collection, and labeling techniques are new areas where the state-of-the-art is changing rapidly, experienced practitioners are scarce, and training materials are not abundant. An engineer needs to know a programming language like Python and be comfortable with accelerator architectures and server environments to get started with training.

Furthermore, to design an ML model that runs on an embedded device, it is also necessary to understand the constraints. These constraints include items like which model layer types the chosen deployment framework supports, how compression techniques like quantization are supported, the read-only/flash and RAM budgets, and what latency is acceptable for inference. The model author will also need extensive knowledge of the sensor’s characteristics (Table~\ref{tab1}) such as capture frequencies, image resolutions, camera lens distortions, and audio preprocessing techniques such as noise reduction, in addition to the particularities of the deployment environment such as the presence of background noise.

The engineer responsible for integrating the model and inference framework into the broader software system needs to understand the usual details of embedded design but also be able to investigate and debug issues around the ML module. If the model is not producing the expected results, is it because the input data is incorrect, is it that feature engineering is failing, is there a bug in the ML framework, or is the output from the model being processed incorrectly?

It is possible to split these responsibilities across multiple engineers so that an embedded systems expert can inform a model author of the constraints and requirements of the target platform. An integration engineer can get debugging help from an ML specialist, but the amount of continuous information transfer needed and the need to have a good enough understanding of both domains to communicate effectively makes this arrangement onerous and risky. In practice, successful deployments have required individual engineers to have or gain proficiency in both ML and embedded software development. People willing and able to embrace this kind of “full-stack” of domain knowledge are rare and hard to train. So lack of suitable engineers has become a critical barrier to the broader deployment of embedded ML applications.

To address these issues, better tooling~\cite{EdgeImpulse:online} that streamlines the development process and makes communicating the required information more accessible will help lower the amount of training required. There is also a need for better educational resources to increase the available pool of talent. Despite numerous existing education and training resources~\cite{warden2019tinyml,TinyMach84:online,plancher2022tinymledu,reddi2021widening,Welcomet0:online,MLOpsfor24:online,Introduc25:online}, these will take time to have an effect and be counterbalanced by growing demand.

\section{Sensor 2.0 Paradigm}
\label{sec:approach}

In this section, we introduce our definition of an ML sensor. We follow through with a few widely used examples of ML sensor implementations to help compare and contrast how such sensors could be realized in the sensor 2.0 paradigm. Next, we present our proposed approach as an initial set of design principles for realizing ML sensors in practice.

\subsection{Definition}

We believe many of the sensor 1.0 paradigm's problems can be solved or reduced by encapsulating ML sensing capabilities in separate hardware components outside the central embedded processor and application code. Like existing primitive types such as pressure, temperature, or accelerometer sensors, ML sensors should expose only a minimal interface through a handful of pins, with no access to the primary address space. We believe this will provide an intuitive way for embedded engineers to easily integrate the newly-emerging advanced capabilities offered by ML without the headaches of the current approach.

We propose the following definition of an ML sensor in simple terms:
\begin{quote}
``\textit{An ML sensor is a self-contained system that utilizes on-device machine learning to extract useful information by observing some complex set of phenomena in the physical world and reports it through a simple interface to a wider system.}''
\end{quote}

This definition is somewhat ambiguous since ``complex'' and ``simple'' are not well-defined. However, it distinguishes these devices from traditional sensors and the current approach to ML integration on embedded systems. Therefore, we believe this definition will still be helpful.

\subsection{ML Sensor Examples}
\label{sec:sensorexamples}

Before we dive deeply into the proposed technical approach, it is helpful to sketch some concrete examples of ML sensors. To this end, we include four examples here, covering different sensor modalities (audio, visual, and telemetry) to show how the proposed concept generalizes.

\subsubsection{Person Detector}
\label{sec:pd}

A common application of computer vision for embedded devices is detecting when a person is nearby~\cite{https://doi.org/10.48550/arxiv.1906.05721}. This output is typically used to wake up a user interface and automatically adjust environmental controls like lights, air conditioning, heating, or doors. A person detector ML sensor would ideally have three external pins, two for power and ground and one signal pin driven high when a person is detected and held high for as long as people are present. It would also have an opening that would be pointed outward at the environment. This information should be all the knowledge an end-user needs to use the sensor.

Internally, the hardware for the person detector ML sensor would be a small camera connected to a microcontroller or digital signal processor. This processor would be running an ML model periodically to determine whether a person was detected in the latest captured frame. If so, it would trigger the single output pin. There may also need to be a near-infrared LED that can illuminate nearby objects when ambient lighting is not bright enough, with a camera capable of capturing those frequencies. If the mounting orientation cannot be guaranteed, an accelerometer could also be present to detect which way the images are up.

\subsubsection{Finger Tap Recognition}

It is possible to use an ML model on accelerometer and gyroscope data to detect finger taps on the casing of a device~\cite{https://doi.org/10.48550/arxiv.2102.09087}. Advanced models can infer the location and even the firmness of the contacts, but this capability depends on knowledge of the shape and material of the case. In many cases, just knowing that a tap or double tap has been detected anywhere would be helpful enough and should be robustly detectable across a wide range of system coverings. An ML sensor using this approach would have three pins, power, ground, and a signal pin driven high for a short period (for example, 200 milliseconds) when a tap is detected. Functionally, this can be used much like a traditional hardware mechanical button, but without requiring any external openings in the case and able to respond to virtual clicks anywhere on the physical device.

Such a sensor could be constructed using an accelerometer, gyroscope, and simple microcontroller. The memory, data rate, and compute requirements of the model are low enough that a comparatively low-cost processor can be used. Because no magnetometer or sensor fusion is needed, a basic inertial measurement unit is sufficient. There may be some complications in designing a model that can work robustly across all the possible variations in cases and mountings, and there are likely to be false triggers. However, these are familiar problems to interface designers who have historically worked with traditional buttons and switches. Therefore, we do not think it will impair its usefulness significantly.

\subsubsection{Voice Command}
\label{sec:voicecommand}

There are a lot of possible approaches to voice interfaces. If one searches the web or dictates text messages, the device needs to understand a wide variety of sentences as faithfully as possible. However, some applications do not require this level of understanding, where an interface only needs to understand a few spoken commands. For example, a light switch might only respond to ``on'', ``off'', and maybe ``dim'' and ``brighten.'' Perhaps surprisingly, this is not a significantly easier problem to solve than the general speech recognition required by more complex interfaces, because there will be many complete words and fragments that sound similar to any chosen target words. These homonyms must be excluded to avoid copious false positives during regular conversation. However, commercial organizations have solved this for particular wake words such as ``Alexa'' or ``Siri,'' though the words themselves are usually chosen carefully to avoid similar sounds used in everyday speech.

An intelligent ML-based voice command sensor could be set up to recognize a small number of words in a particular language, either some collection of broadly-applicable commands such as ``on'', ``off'', ``stop'', ``play'', and so on, or sets that are more specific to a particular class of use cases. If the choices are ``on'' and ``off'', the sensor can follow other examples of ground, power, and a single signal set high when ``on'' is heard and driven low for ``off''. For more complex word collections, a serial protocol such as I\textsuperscript{2}C will be required so that the sensor can send small packets that indicate what command was recognized to the central processor or system-on-chip in the system.

The implementation for this ML sensor in its simplest form will consist of a microphone and a microcontroller, running an ML model to detect the chosen words. A camera could also be used to increase the overall accuracy, both by integrating gaze detection to ensure the command is aimed at the device and by using video to improve recognition in noisy environments~\cite{ephrat2018looking}.

\subsubsection{Text Reader}
\label{sec:textreader}

Seven-segment numerical displays have been common on equipment since the 1970’s. They were originally designed for human readability, but ML models can now interpret them with high accuracy. Teams involved in areas such as agriculture and water distribution~\cite{chinawater, digitizer} have been working on applications that apply such models to retrofit existing hardware systems that use these displays to output information to nearby people, to also feed into secondary systems for further processing. For example, an initially-offline soil moisture analyzer can be upgraded to transmit values periodically over the internet for remote monitoring, using a small IoT board connected to a camera.

Since this is expected to be a common use case, an integrated sensor component that handles reading digital displays and can pass the numerical value to another module could be helpful. The best interface design is not as evident as some other examples. However, it could consist of a serial protocol like I$^2$C that transmits two thirty-two-bit values periodically, the first being the signed whole number portion of the read value, and the second being the fractional part. These should be transmitted in binary-coded decimal format to avoid any conversion loss from the original displayed values. This would accommodate up to eight digits on either side of the decimal point.

The implementation for this ML sensor would rely on a small camera connected to a microcontroller running a digit recognition model. The goal would be to make it as general and easy to use as possible, so the model would ideally be able to find the string of digits no matter the orientation or position of the display within the camera frame. It may be necessary to include a small display on the back of the sensor to help installers position it correctly if there are limitations on the model. There may also need to be some kind of lighting for segment display types that do not self-illuminate. If mechanical numerics, gauges, dials~\cite{lilz}, or even arbitrary text on a screen can be read, it may be possible to extend the sensor concept to handle these as inputs.


\subsection{Design Principles}

The examples of ML sensors we described support various use cases and have many common properties. Therefore, we propose that a valuable and feasible ML sensor is defined by the following externally visible attributes.

\subsubsection{Slim Interface}

The ML sensors' hardware and software must have simple and slim interfaces. The internal implementations of most ML sensors will involve a general-purpose processor or a digital signal processor (DSP). It may be tempting to expose the programmability available through a complex API to enable features like model updating or other customizations on the software front. We believe this would be a mistake. Minimizing configurability allows testing, security, and privacy evaluations to be performed with higher confidence in their results. It is also helpful for application builders to reduce the knowledge required to integrate the sensors into their systems. As we discussed in Section~\ref{sec:sensorexamples}, most ML sensors should have a straightforward hardware interface with three pins on the hardware front. If additional functionality or flexibility is needed, serial protocols like I2C or SPI could be readily adopted.

\subsubsection{Reusability}

All ML sensors should be usable for a variety of different applications. It is more effort to build a self-contained component than it would be to integrate ML software and models into a single embedded system using the current approach. Thus, the investment must provide benefits over multiple applications to make commercial sense. It is much like the trade-off between writing a software library or just writing ad-hoc code to solve a problem. The former encourages code reuse and robust testing and is just a better software engineering practice, while the latter gets the job done in the short term but proves harder in the long term for maintenance. There should be evidence that any solution will be used by more than one product before a decision is made to build it more generically.

\subsubsection{Composability}
\label{sec:compose}

ML sensors ought to be designed such that they are composable. One should be able to treat individual ML sensors as building blocks that can be pieced together with other ML sensors to realize more sophisticated ML sensors.

Consider, for instance, the voice command sensor we discussed  (Section~\ref{sec:voicecommand}). Wake words such as ``Siri,'' ``Alexa,'' or ``Hey Google'' are becoming ubiquitous interfaces for everyday appliances. As the number of voice-controlled devices in an environment grows, it becomes cumbersome to know or specify which device a user is talking to as part of a command such as ``Siri, living room lights off.'' 

We believe that gaze can become a functional building block for more intuitive interfaces in such a scenario. Eye contact is a vital communication cue for both people and animals. It gives us confidence that any speech or non-verbal signals are directed at us rather than anyone else. However, it is still not common to use this cue for our interactions with machines, though there are some signs of adoption, such as phones hiding or showing their screens depending on whether they detect a user looking at them. We believe that using gaze to indicate which device the user is talking to when providing the command, such as looking at a light switch and saying ``off'' is a better method of realizing a voice command sensor so that all devices do not turn on.

To support the more intelligent version of the basic voice command sensor, we propose a gaze sensor with power, ground, and a signal pin driven high when any person looks directly at the component. It will require an opening for the camera, and its internal implementation will be similar to the person detector sensor. When the gaze detector signal pin goes high, it can be coupled with the voice command sensor's signal pin to determine if the user was giving a command to the device under consideration.
\newpage
This example shows that it is possible to build ML sensors that can sense more complex phenomena using more basic ML sensors. Enabling and supporting such composability by design will allow each ML sensor to be robustly tested and audited before being integrated into a wider system.

\subsubsection{Calibration}

Thus far, we have focused on widely-applicable problems where a single ML sensor model can be reused across many different products. However, there is another class of problems where the ML sensor requirements are similar to those solved by a well-known model, but the recognition targets themselves are domain-specific.

For example, a farm might want to recognize rodents instead of people. ML can usually produce reasonable solutions to these shifts in domains with comparatively small amounts of data using transfer learning~\cite{weiss2016survey,zhuang2020comprehensive,yosinski2014transferable}. This technique entails taking a model trained on a classic problem with an extensive data set and fine-tuning the parameters by running training on the new classes, possibly only updating a few of the later layers in the network using transfer learning.

Fine-tuning the top layers of an ML sensor model to its domain-specific task is akin to calibrating a traditional hardware sensor to its environmental characteristics. Transfer learning forms the basis of many ML cloud services. We believe it would be a valuable extension to preset ML sensors in many cases. Transfer learning could be supported by uploading customized models to an ML sensor, either by the sensor provider or by the system integrator. If the system integrator needs access, this may involve widening the thin API we have discussed, so an approach that occurs during sensor manufacturing might be preferable. The risk should be manageable since this can be implemented by uploading only model parameters with no executable code modified.


\subsubsection{Connectivity}
\label{sec:connectivity}

WiFi, cellular, LPWAN, and other network connections are standard on the microcontrollers likely to be used for ML sensor implementations. We believe that including networking capability within these ML sensors themselves is a bad idea. Any network connection can be seen as a pretty wide API and dramatically increases the attack surface for security and privacy exploits. It should be possible for third parties to audit ML sensors to ensure with a high level of confidence that they cannot be misused to spy on people, and networking makes this very hard to do.

\subsubsection{Updatability}

Another particular case of keeping the interface minimal is removing the ability to update the software used by the internal implementation of the ML sensor. This approach may seem perverse since the capability is likely to be supported by the sensor's processor. However, ensuring the behavior of a sensor cannot be changed after it has been deployed is vital if users are going to have confidence in the results of any privacy guarantees, which we believe take precedence.

However, the inability to update the ML model within the sensor once deployed means that it may also be challenging to resolve issues that arise due to the lack of connectivity, as discussed in Section~\ref{sec:connectivity}, precluding the possibility of addressing concept drift~\cite{widmer1996learning}. Concept drift occurs when the statistical properties of the target variable, which the model is attempting to predict, alter in unanticipated ways over time. This drift is problematic because the accuracy of predictions decreases as time passes and can occur suddenly, gradually, incrementally, and it may even recur over time. Therefore, it is generally essential to detect.

\subsection{Implementation Requirements}

If ML sensors gain widespread adoption, they will need to be attractive to embedded system designers. So their implementations will need to conform to constraints.

\subsubsection{Standardized Interchangeable Parts}

Providing ML capabilities behind a simple abstract interface will only simplify system building if the abstraction does not leak. Different implementations of a given class of ML sensor will naturally have different performance characteristics depending on the hardware, software, and models used internally. However, we should aim for the interface itself to remain constant. For example, person detector sensors (Section~\ref{sec:pd}) from different sources might differ in their sensitivity, performance in different lighting conditions, and update latency, but there is no reason that they should expose their signals in different ways since they are fundamentally trying to report on the same basic information. Suppose the ML sensor architects can agree on interface standards for common types. In that case, it will increase the attractiveness of these components for system developers and avoid unnecessary lock-in to particular suppliers. We discuss this standardization further in ML sensor datasheets (Section~\ref{sec:datasheet}).

\subsubsection{Conformance Testing}

One of the challenges for embedded product developers is that they have no good way to predict how well an ML model will work in a deployed system ahead of time because of all the previously discussed challenges. Having well-defined testing protocols for types of ML sensors would ease this problem. For example, a person detector sensor might be placed in a room with a calibrated monitor at a set distance displaying a series of labeled test images and the percentage of true and false positives measured. Defining these tests may be an involved process; for example, this scenario does not consider lighting or near-infrared capabilities, but even imperfect tests would help system builders compare alternative components and plan for realistic performance.

\subsubsection{Auditability}
\label{audit}

The most concerning aspect of the status quo for ML in embedded systems is how it encourages the broad deployment of cameras and microphones with no way for end-users or even system integrators to check that they are not compromising their users’ privacy by enabling recording. Centralizing the risk within a single component and minimizing the interface to the rest of the system is necessary to enable the kind of third-party privacy audits that would offer more confidence for users, but it is not sufficient. 

We should also expect ML sensor providers to design their implementations to make it easier for third parties to evaluate important properties. For example, using an off-the-shelf microcontroller with no WiFi, Bluetooth, or cellular capability should make it straightforward for someone to verify that networking is impossible. Source code inspection may also be necessary to achieve high confidence. However, because the principles have minimized the attack surface we covered earlier, and the software is a self-contained module running in a separate address space from the rest of the system, it ought to take less time than would be required for a regular software library integrated into a device.

\subsection{Potential Pitfalls and Challenges}

Thus far, we have primarily focused on the advantages of the proposed approach. To provide a balanced point of view, we also outline the potential pitfalls and challenges of our proposed approach to realize ML sensors' capabilities. 

\subsubsection{Cost}

Embedded systems often have very tight cost constraints, and essential design decisions can be made based on differences of just a few cents. In this context, expecting system builders to pay for the extra microcontroller within an ML sensor could seem unrealistic. We believe that this will not prove insurmountable for several reasons. First, there are applications where the added user value and reduced software development costs (or non-recurring engineering (NRE)) will outweigh the extra expense, so we expect initial customers even for early versions of the sensors. 

Second, we expect the cost of sensors with integrated compute capability to drop rapidly over time. Increasingly, we already see camera modules, such as Sony’s Spresense~\cite{Overview16:online} or Himax’s Wise-eye~\cite{HIMXAnno37:online} devices emerging, and we expect that trend to continue. These compute-enabled sensors are designed to run data processing in a separate module from the processor controlling the primary system, so they are well suited to our suggested design approach. With the software development costs amortized over many different deployed systems, we hope the economics of scale will favor picking ML sensors in many cases.

\subsubsection{Power}
 
The TinyML movement targets devices capable of running ML with less than one milliwatt of average power usage to enable battery-powered and energy harvesting applications. By adding a separate processor to deal with sensor data analysis rather than integrating it into software running on an existing microcontroller, the ML sensor approach may seem like a step backward in power efficiency, and we do indeed expect that such systems will use more energy to operate in the near term. Early adopters are likely to be cases where reducing power usage to single-digit milliwatt levels is not the highest priority, such as devices connected to mains electricity. However, we believe that reducing the power usage significantly over time should be possible, especially as the sensor processor hardware can be specialized for a single workload. In many cases, the signal from the model run on the sensor data can wake up the general-purpose processor, allowing it to hibernate in a shallow power mode the rest of the time. If the ML processing is run as part of a general software stack on the main microprocessor, it often requires that all processor components are powered up, even if the ML does not use them. There is often little choice about waking different modules like the memory that can take much energy even if unused, and sensor data processing typically requires a cadence of running multiple times a second, more frequently than most other tasks. Breaking the sensor work out into another physical component makes it easier to separate just the parts needed into their own power domain, hopefully reducing overall energy consumption.

\subsubsection{Universality}

One of the underlying assumptions of the ML sensors concept is that there are problems that occur across many different applications that can be solved with the same model. The examples listed earlier in Section~\ref{sec:sensorexamples} assume that tasks such as detecting when a person is present are helpful to many different products and can be adequately addressed by a single model. There is some evidence for this in cloud ML APIs, which offer capabilities like speech or face detection~\cite{Detectfa79:online} designed to work without knowledge of the end-use case. However, these models typically require more time and resources to train than application-specific networks. They may also fail to perform as well as more specialized models in situations not present in their training data. For example, a person detector may have been trained with the implicit assumption that the input images are shot horizontally, and images from a  sensor mounted on the ceiling looking downwards showing the tops of people’s heads may not be detected accurately. Similar problems may occur if the backgrounds, environmental conditions like smoke or fog, or clothing vary from those present in training data.

We believe it is possible to develop models capable of working across a wide-enough domain of problems to be usable for enough applications. Unfortunately, this question of universality can only be honestly answered through engineering and then observing the results.
\section{Datasheets for ML Sensors}
\label{sec:datasheet}

A datasheet is a document that specifies the features and characteristics of a product. Traditional commercially available sensors are sold with datasheets that outline their hardware and operating characteristics. These characteristics include parameters such as power consumption, operating temperature, and the specifications of the sensing application (e.g., detection limit, measurement frequency). 

Datasheets are necessary as they allow developers to determine whether the specification of the sensor is suitable for their particular application. Additionally, they act as a performance reference that can be independently assessed for quality assurance purposes. Some sensors may be used in performance-critical workflows and, as such, require rigorous evaluation before being implemented into workflows. 

ML sensors will require something analogous to the datasheet of traditional sensors. However, this new datasheet will need to document not only the information included on a traditional sensor datasheet but will also need to capture the characteristics of the machine learning model, the dataset(s) used to train and test the model, and end-to-end metrics that capture the impact of environmental characteristics on the accuracy of the device. An informed user can determine whether the device is suitable for their particular application.

There has been a growing focus on producing new datasheets for ML in recent years. These range from datasheets for datasets~\cite{gebru2018datasheets} to nutrition labels~\cite{holland2018dataset} which provides a standard label for understanding a dataset's characteristics. Even specialized application areas of ML, such as medical prognosis and diagnosis, are starting to build their own custom datasheets out of the need for better transparency~\cite{sendak2020presenting}.

\subsection{Example Datasheet} 

\begin{figure*}[hbtp]
    \centering
    \includegraphics[width=.9\textwidth]{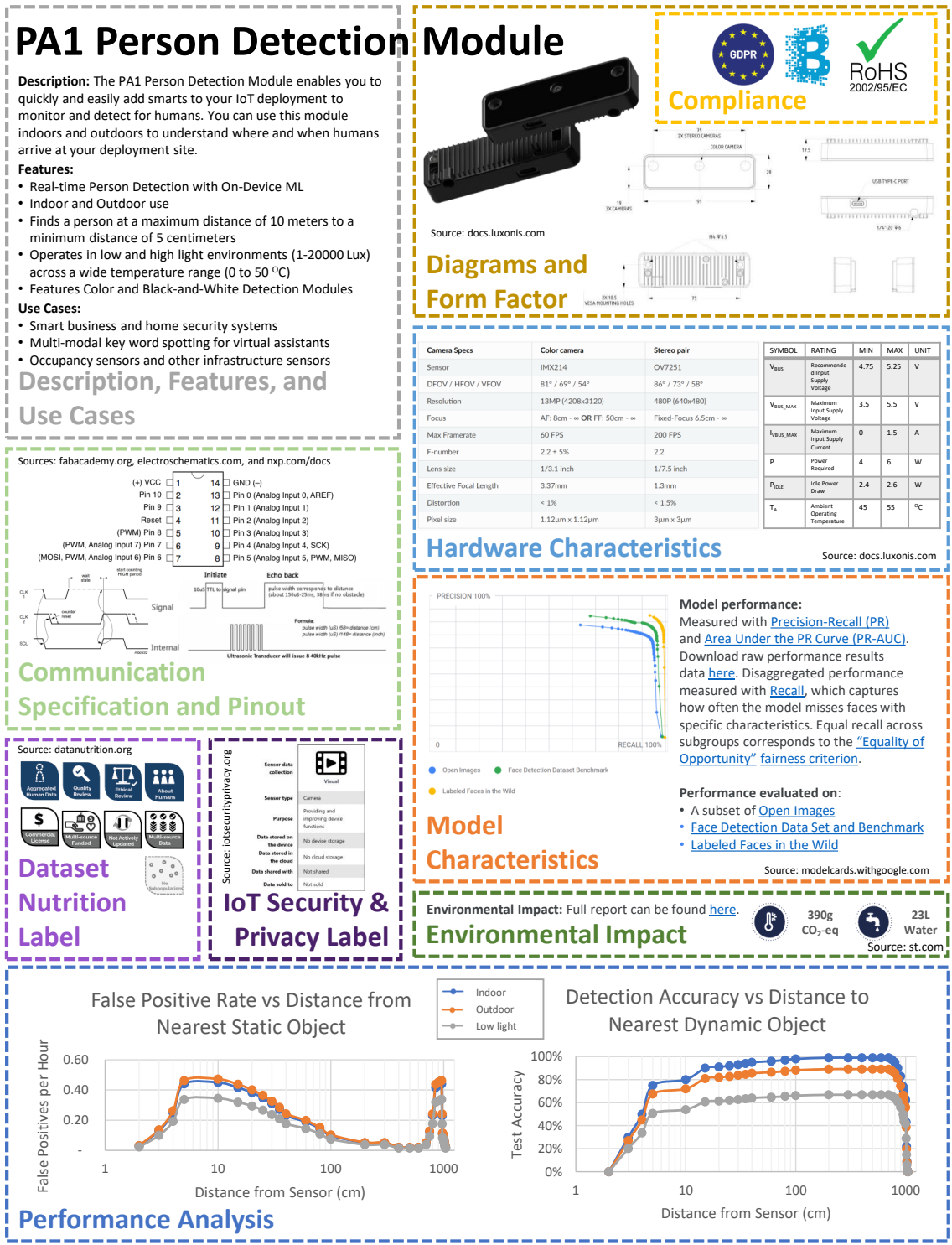}
    \caption{Illustrative example datasheet highlighting the various sections described in Section~\ref{sec:datasheet}. On the top, we have the items currently found in standard datasheets: the description, features, use cases, diagrams and form factor, hardware characteristics, and communication specification and pinout. On the bottom, we have the new items that need to be included in an ML sensor datasheet: the ML model characteristics, dataset nutrition label, environmental impact analysis, and end-to-end performance analysis. While we compressed this datasheet into a one-page illustrative example by combining features and data from a mixture of sources, on a real datasheet, we assume each of these sections would be longer and include additional explanatory text to increase the transparency of the device to end-users. Interested users can find the most up-to-date version of the datasheet online at \url{https://github.com/harvard-edge/ML-Sensors}.}
    \label{fig:example_datasheet}
\end{figure*}

An illustrative example of a datasheet for an ML sensor is shown in Figure~\ref{fig:example_datasheet}. It focuses on a person detector sensor (Section~\ref{sec:pd}), which we envisage as a broadly applicable device. This section breaks down the proposed structure of the datasheet in more detail. As stated above, our proposed datasheet summarizes not only aspects of the sensor hardware but also of the machine learning model, the datasets used to train and test the model, the security, privacy, and environmental impacts of the model, and end-to-end application-specific performance characteristics; this type of information is unique to the ML-based sensors.


\subsubsection{Description, Features, and Use Cases}

This section is essentially unchanged from that seen in a traditional sensor datasheet. A high-level description of the device is provided with some of its most essential characteristics, such as the underlying processor, communication protocols, and sensors utilized in the platform, summarized in one to two paragraphs. Features are a bullet-pointed list of the most salient aspects of the device. For a person detector, this might include the operating range and luminosity, information about the embedded camera, the device's size, the form of output, and other characteristics. Use cases describe typical applications where the device might be used. For a person detector, everyday use cases might include intelligent lighting or security systems or as part of a cascade system that activates additional processes such as person identification or counting individuals in a room.

\subsubsection{Compliance} 

The compliance section will often only include a few statements or marks that demonstrate compliance with some form of government or inter-governmental regulation or an industry-specific expectation. Since ML sensors dovetail ML with traditional embedded hardware, compliance factors relevant to both industries will need to be highlighted. The presence or absence of specific compliance marks may allow or prohibit the device from being sold or used in specific countries or regions. Examples of this might include being lead-free~\cite{2022RoHS41:online,EURLex3264:online} or GDPR-compliant~\cite{voigt2017eu}.

\subsubsection{Model Characteristics} 
\label{sec:modelchar}

This section focuses on the characteristics of the machine learning model integrated within the device. Similar to a model card~\cite{mitchell2019model} or an ML-model fact-sheet~\cite{arnold2019factsheets}, an ML sensor's characteristics may include information regarding the training and test data set, such as its size and whether it is open-source or not, the data modality, as well as information regarding the model architecture. Additional information on the datasets used to train and test the model may also be provided, allowing users to understand better the device's potential biases and, therefore, better judge its applicability to their application. Information could also be provided detailing whether some form of third-party entity has validated this performance.

\subsubsection{Dataset Nutrition Labels}

Building on the concept of nutrition labels~\cite{holland2018dataset,TheDataN2:online} and datasheets for datasets~\cite{gebru2018datasheets}, the ML sensor datasheet will need to include information on broader impacts of the ML sensors and the ML model upon which it is based. Data nutrition focuses on issues associated with the ML sensor's model training data. It helps us understand, for instance, whether the underlying training data meets the licensing agreements that permit the use of the model and whether that data was subject to ethical reviews. 

\subsubsection{IoT Security \& Privacy Labels}

As we develop more consumer-facing devices that rely on machine learning (ML) sensors, there is a need to assist consumers and system implementers in making informed purchasing decisions. To this end, it is crucial to ensure that ML sensor manufacturers disclose their privacy and security policies. In this regard, we suggest building upon CMU's IoT Security \& Privacy Label~\cite{https://doi.org/10.48550/arxiv.2002.04631,megas2021establishing} efforts, which include information on the privacy and security practices of the smart device, such as the type of data the device collects and whether or not it receives automatic security updates.

This section is where we can highlight what data is available to the rest of the system and what is not, along with the certification of any external auditing process performed. For example, we could mention that the module uses a camera but only outputs whether a person is present or not. This information could also be shared on consumer-facing product descriptions to help end-users understand what privacy safeguards are included.

\subsubsection{Environmental Impact}

The ML sensor datasheet should also include information on the environmental impacts of training and running the machine learning model used in the sensor. Including the ML sensor's footprint in datasheets helps bring awareness to the large environmental impact that can be imposed by the machine learning lifecycle~\cite{wu2021sustainable}. Moreover, while the footprint of a single low-cost, low-power microcontroller is much smaller than that of cloud ML solutions~\cite{stmicro}, it is not negligible. Thus, given ML sensors' projected large-scale deployment, the environmental impact must be considered. 

\subsubsection{End-to-end Performance Analysis} 

While the information on the model and data nutrition characteristics are necessary for an ML sensor, it is not sufficient to inform potential consumers. Some type of information must be available to the consumer to assess how the device's performance might differ based on its specific application, specifically in real-world characteristics. For a person detector sensor, many aspects could be relevant to its performance, such as the lighting environment and operating distance. Thus, this section aims to provide that information to potential purchasers by outlining how environmental factors that might naturally be anticipated in the data could be incorporated into the end-to-end performance of the device. This way, there is some characterization of how a difference in the data distribution compared to the training data might alter model performance and impact the device's applicability for a user's target application.

\subsubsection{Diagrams and Form Factor} 

In addition to the unique characteristics that ML brings to sensor 2.0, when buying any sensor, it is often helpful to understand its size and shape for design and integration purposes. This information might inform a purchaser whether the device is suitable for their application, or they may choose to build around the device specification. Often, CAD models of sensors are available online so that product designers can integrate these into their 3D models.

\subsubsection{Hardware Characteristics} 

Hardware characteristics focus on the traditional aspects of the sensor, such as its operating temperature, power consumption, input voltage, and electrostatic discharge rating. These electrical characteristics highlight the low-level, hardware-relevant aspects of the system, separating it from the ML-relevant aspects of the system, which are examined at a higher level of abstraction in the model characteristics section of the datasheet as described in Section~\ref{sec:modelchar}.

\subsubsection{Communication Specification and Pinout} 

Purchasers will use the datasheet as reference documentation for the sensor during both design and implementation, meaning that information regarding its pin layout and communication protocols is necessary to outline. This section will show the user how to interact and use the sensor at a hardware level to power the sensor, provide data to the sensor, and extract and interpret the output.
\section{Ecosystem Development}
\label{sec:ecosystem}

The ecosystem surrounding ML sensors can be imagined as either closed source, open-source, or hybrid approach. There are pros and cons related to each of these approaches. 
This section presents our view of why an open ecosystem around ML sensors is crucial to develop.  Similar to how open-source development for both ML and other software has radically transformed the software ecosystem, we believe an open-source ecosystem around ML sensors can radically accelerate the pace and innovation of ML sensors. 

\subsection{Closed versus Open Source}

Fostering an open ecosystem will allow the development of more flexible and innovative solutions than a closed ecosystem, helping to maximize the impact of ML sensors from both commercial and research perspectives. The open nature of collaboration will also stimulate community-built solutions that will likely be superior to any created by a sole developer. The transparency of the solution will build trust among users. Open source means it is faster to identify issues and file bug reports, which results in quicker fixes. Given that one of the major hurdles facing embedded ML is the need for full-stack expertise (Section~\ref{sec:challenges}), an open-source community will be able to build ML sensors collaboratively.

However, achieving the vision of an open ecosystem while still allowing sensor developers to maintain their intellectual property, such as the trained model weights, will likely prove challenging. Closed source solutions will protect intellectual property but will potentially suffer from monopolies that naturally lend themselves to a lack of interoperability caused by ML sensor heterogeneity and potentially reduced trust due to the black-box nature of the ML sensors.

To address this issue, a hybrid approach may prove optimal. Allowing an ML sensor's architecture to be open source while keeping the trained model proprietary could be a fair compromise. Also, a validation set could be provided to users to allow probing of the model output and performance while keeping the training data used for the algorithm and explicit model architecture undisclosed. This approach benefits partial transparency while still providing an economic incentive for companies to develop innovative solutions. In the case of ML sensors that are applicable across many different use cases, the community may create entirely open source solutions. At the same time, more specialized applications remain in the wheelhouse of bespoke developers.

\subsection{Large, Public Sensor Datasets}

One of the critical limitations of growing the ML sensor ecosystem is the lack of public sensor datasets. Since ML systems require large and high-quality datasets to develop robust models, an equivalent set of datasets is required for ML sensors. Computer vision specialists have ImageNet~\cite{https://doi.org/10.48550/arxiv.1409.0575} at their disposal, while natural language researchers have SQuAD~\cite{https://doi.org/10.48550/arxiv.1606.05250}. A big question for the ML sensor community will be \textit{``What is the ImageNet of ML sensors?''}

Large public datasets are needed to accelerate ML sensors' design, implementation, and deployment. Preliminary analysis of major technical organizations shows the widespread use and adoption of open data sets to establish baselines~\cite{https://doi.org/10.48550/arxiv.2102.11447}. Open data sets are vital to accelerating ML innovation for everyone, but such resources remain scarce across the ML ecosystem. Thus, a key question facing the ML sensors community is if dataset creation can be accelerated akin to the rapid development of open-source software.

To this end, we recommend creating a ``Data Commons'' aimed at making sensor data of all forms publicly accessible to the community under the appropriate licensing terms. A crucial requirement here is to follow a standard sensor data-file format, perhaps similar to the Open Data Format (O-DF) standard designed for IoT devices~\cite{OpenData54:online}.

\subsection{Benchmarks and Toolchains}

To foster a rich and open ML sensor ecosystem, we need full-stack frameworks that are publicly accessible to everyone. The full-stack includes software frameworks for training and running ML models efficiently, as well as having a flexible hardware infrastructure for accelerating ML operations inside the ML sensor's custom processor.

On the hardware front, given that ML sensors will be task-specific, it makes sense to develop application-specific processors (ASIPs)~\cite{jain2001asip}. Building a tiny, custom ASIP for the ML sensor itself (see Figure~\ref{fig:ml_sensor_v2}) with an instruction set architecture (ISA) tailored for the specific ML model provides greater energy efficiency than an off-the-shelf processor. The RISC-V~\cite{asanovic2014instruction} hardware ecosystem is well poised for enabling custom hardware acceleration for ML sensors.

Software frameworks such as Gemini~\cite{genc2021gemmini} and CFU Playground~\cite{prakash2022cfu} are needed to enable the development of open-source hardware. As these are full-stack software frameworks capable of running the software and hardware, they naturally lend themselves to hardware and software co-design. As ML sensors are bespoke, co-design can offer a significant improvements for low-power, always-on ML.

Such an open hardware and software ecosystem will lead to the proliferation of ML sensor solutions. We will need new benchmarks and associated metrics to assess the various emerging solutions. Benchmarks such as MLPerf Tiny~\cite{banbury2021mlperf} can help streamline apples-to-apples comparisons of different sensors. However, these existing benchmarks will need to evolve to meet the needs of ML sensors. For instance, for a person detector sensor, in addition to capturing traditional metrics such as detection accuracy, we will need to capture additional metrics such as the minimum and maximum distance at which the ML sensor will work reliably. 

\subsection{Sensor Libraries}

As the ecosystem matures, the self-contained nature of ML sensors will result in modular devices built for a singular purpose. Naturally, this approach lends itself well to the potential for composability (Section~\ref{sec:compose}), allowing multiple ML sensor devices, as well as traditional sensor devices, to be combined as part of wider system architectures that can achieve more specialized goals. 

Analogous to how open source libraries are created, the ecosystem may work together to determine what devices are most relevant to create, as well as an approach to mix-and-match devices and communication protocols between them that can be used to augment individual outputs. The composability of ML sensors alongside traditional sensors will provide a new and powerful toolkit for developing more specialized ML-enabled hardware devices.

\section{Ethical Considerations}
\label{sec:ethics}

While we anticipate the impact of ML sensors will be significant and net-positive, inevitably, there may be some adverse impacts. The associated risks of these negative effects can be minimized by limiting, for example, connectivity and updatability (Section~\ref{sec:approach}). However, there is no fool-proof way to guarantee that such methods can eliminate potentially harmful applications. Consequently, care and consideration must be shown to discern how potential harm can be mitigated in the context of sensor 2.0.

\subsection{Concerns}

The growing deployment of online devices running autonomously for extended periods opens up many opportunities for malicious actors~\cite{gray2016always}. Our proposal aims to ensure that as many devices as possible have built-in safeguards against accessing personal data like recordings in their default state. However, there are plenty of other possible harms that the proliferation of ML sensors is likely to enable, and some that may be made worse by advances in on-device ML. Existing ethical challenges in the realm of traditional ML remain in the sensor 2.0 paradigm, but additional considerations are necessary to determine how running ML locally may cause specific problems.

There is a growing realization that AI has the potential to revolutionize warfare. ML is already being used to develop automatic target recognition systems~\cite{DeepLear34:online}. On-device ML sensing may push these dangerous capabilities one step further. An example use case of on-device ML is in learning-based autonomous weapons systems (AWS)~\cite{hua2019machine}. A missile could guide itself towards a target using ML sensors. Alternatively, it is also possible to engineer a landmine or improvised explosive device (IED) that triggers when a person is nearby, or worse, when a person of a particular demographic comes close.

Even outside military applications, organizations could develop generalized ML sensors that distinguish between ethnicities or gender presentations and use these to discriminate on a massive scale. Another potentially harmful application could be an audio-based sensor that triggers when particular words like ``democracy,'' ``abortion,'' or ``human rights'' are spoken nearby, alerting repressive authorities to potential dissidents. Applications like these are possible even without our proposed approach but could be deployed far more efficiently and widely thanks to their properties.

Last but not least, it is also possible that malicious or careless manufacturers take advantage of any increased public trust in devices offering privacy through on-device processing to promote hardware that is not as secure. For example, their microphone might not be genuinely segregated from the main microcontroller and be vulnerable to access. 


\subsection{Possible Solutions}

Because many of the issues with our approach are scaled-up versions of general problems created by ML being deployed, many of the same tactics can be used to reduce likely harms. All ML sensors should have a publicly available datasheet (Section~\ref{sec:datasheet}) that discusses the essential properties of the system so that product integrators and end-users can be aware of limitations and quirks. We also believe that the possibility of having a third-party service audit the privacy claims is an advantage of our approach, and recognized standards and a certification process should be developed in collaboration with organizations like Underwriters Laboratories~\cite{ULEmpowe80:online} or Consumers' Union~\cite{ProductR18:online}. 

Dealing with determined malicious actors is a lot more challenging. Public availability makes the prospect of, for example, terrorists using person detectors to trigger improvised explosive devices very concerning and hard to combat. The hardware modules will be available to buy freely and cheaply from self-service component distributors like DigiKey or Mouser, so screening prospective customers will not be possible. Even if we identify and block problematic initial purchasers, they will be built into everyday devices that are commonly available and modular. Thus, malicious actors extracting them from existing products is a genuine concern. Given this likely proliferation, we will need to rely on similarly limited approaches for other electronic components with dual uses, primarily investigation and enforcement.
\section{Related Work}
\label{sec:related}

The landscape of ML sensors is rapidly growing, with novel applications emerging regularly, and, to make them run efficiently, the hardware and software ecosystem is also booming. We summarize several compelling use cases and novel system designs and discuss how these innovations are still applicable in the context of our sensor 2.0 paradigm.

\subsection{Application Areas} 

Systems that use sensors and machine learning are becoming rapidly adopted into the ecosystem. Examples range from uses in the home or office for keyword spotting~\cite{zhang2017hello} using audio sensors and person detection~\cite{banbury2020benchmarking} using vision sensors to industrial settings for predictive maintenance using audio, motor bearing, or IMU sensor data~\cite{banbury2020benchmarking}. In addition to these classic tinyML examples, there are also applications deployed in wildlife settings using hydrophones (underwater microphones) to prevent collisions with whales and ship~\cite{johnson2020}. More forward-looking applications also exist. For example, tiny unmanned aerial vehicles with embedded intelligence have been deployed to improve pesticide application in agriculture~\cite{king2017technology}.

ML sensors will be helpful across all these application areas and more. We believe that ML sensors will accelerate the adoption of TinyML into these and many other application areas because the sensor 2.0 paradigm lowers the barrier to adoption. Since ML sensors are the physical embodiment of ML with the addition of sense, it reduces the friction commonly found in the ecosystem, which is that developers and users struggle to cope with understanding ML technology.

\subsection{Hardware Technologies}

Several hardware advancements on the horizon promise to increase the efficiency of ML sensors dramatically. Hardware such as 
flexible electronics~\cite{biggs2021natively},
compute-in-sensor~\cite{moin2021wearable},
compute-in-memory~\cite{chang2021energy}, 
analog computing~\cite{draghici2000neural},
and neuromorphic computing hardware~\cite{bouvier2019spiking} can dramatically increase the energy efficiency of ML at the edge. Due to substantially different memory access and synchronicity approaches, they can be complex to implement in conjunction with a traditional application processor. We believe these will be resolved as use cases become obvious.

The sensor 2.0 paradigm abstracts the hardware differences behind a simple asynchronous interface, thereby enabling the hardware design space of ML sensors to be unconstrained by the application. We can leverage the hardware technologies to build highly specialized hardware for particular ML sensor tasks. The investment required to build such an application-specific integrated circuit will undoubtedly be more significant than that needed to develop software. Nevertheless, if the potential market is large enough, this approach could provide a good return on investment.

\subsection{Model Optimization} 
Recent advancements in efficient model design~\cite{banbury2021micronets}, quantization~\cite{park2018value}, pruning~\cite{blalock2020state}, knowledge distillation~\cite{gou2021knowledge}, and compression~\cite{kim2019efficient} have made it possible to deploy modern ML models to severely resource-constrained devices, enabling the sensor 2.0 paradigm.
However, many of these optimization techniques are typically ignored in traditional edge and tiny ML applications due to the technical expertise needed to take them into production robustly.
\newpage
Under the new paradigm, more aggressive model optimizations are possible since the additional complexity is borne by the ML sensor designer and not each application developer, therefore making the approach more scalable.
Furthermore, the model can be co-designed explicitly with the physical sensor(s), increasing efficiency. This space is largely unexplored, leaving room for future work and new research.

\subsection{Security} 
Previous work has aimed to protect user privacy in sensor applications by encrypting the data in place or in transit~\cite{wang2014performance}. This approach assumes the application developer is trusted and is prone to man-in-the-middle attacks~\cite{gou2013construction}.
Embedded ML aims to preserve privacy by keeping user data on the device, but this assumes the application-level code is trusted.
Previous work has proposed running ML models in secure enclaves to prevent untrusted users from accessing raw sensor data~\cite{prabhu2020privacy}. However, privacy is difficult to ensure without hardware isolation, even with secure enclaves.

As discussed in sections \ref{sec:connectivity} and \ref{audit}, the sensor 2.0 paradigm isolates the user's sensor data and exposes only high-level information to the application. This approach preserves the user's privacy from hackers and potential misuse by an otherwise trusted party.
In ML sensors, the users interpret the exposed high-level information, making it easier to audit what information their devices expose.

\section{Conclusion}
\label{sec:conclusion}

We believe the modular sensor 2.0 approach to system design proposed herein for ML plus sensor capabilities can offer many advantages to both manufacturers and end-users. Product builders benefit from having the complexity of an ML implementation hidden behind a simple interface. Consumers gain new assurances about the privacy of sensitive information. Unanswered questions remain regarding the economics, power consumption, and the generality of singular solutions to multiple contexts, but these can be addressed through future prototyping and research. We think a new approach is needed based on our experience deploying embedded ML solutions. We hope that our proposal forms the basis for a dialogue on how this can be achieved. 

We encourage interested parties to visit \url{mlsensors.org} where we hope to foster a community around ML sensors. In this article, we have outlined several challenges and put forth many suggestions and recommendations for how one might go about addressing the challenges. But we believe that it is challenging for any single organization to answer all the questions. Nor is it possible for an organization to tackle all the concerns in a fair and representative manner that meets the embedded ecosystem's scale and broad needs. To this end, there is a clear need for community-level engagement.



\bibliographystyle{mlsys2022}
\bibliography{refs}



\end{document}